# Stroke Disease Classification through Machine Learning Models by Feature Selection Techniques

**Mahade Hasan**[1,3], **Farhana Yasmin**[2,3], **Xue Yu**[1]

[1]School of Software, Nanjing University of Information Science and Technology, Nanjing, China

[2]Department of Computer Science and Technology, Nanjing University of Information Science and Technology, Nanjing, China

[3]Those authors contributed equally to this study.

Corresponding Author: Mehede Hasan (mhasan@nuist.edu.cn)

## Abstract

Heart disease remains a leading cause of mortality and morbidity worldwide, necessitating the development of accurate and reliable predictive models to facilitate early detection and intervention. While state of the art work has focused on various machine learning approaches for predicting heart disease, but they could not able to achieve remarkable accuracy. In response to this need, we applied nine machine learning algorithms XGBoost, logistic regression, decision tree, random forest, k-nearest neighbors (KNN), support vector machine (SVM), gaussian naïve bayes (NB gaussian), adaptive boosting, and linear regression to predict heart disease based on a range of physiological indicators. Our approach involved feature selection techniques to identify the most relevant predictors, aimed at refining the models to enhance both performance and interpretability. The models were trained, incorporating processes such as grid search hyperparameter tuning, and cross-validation to minimize overfitting. Additionally, we have developed a novel voting system with feature selection techniques to advance heart disease classification. Furthermore, we have evaluated the models using key performance metrics including accuracy, precision, recall, F1-score, and the area under the receiver operating characteristic curve (ROC AUC). Among the models, XGBoost demonstrated exceptional performance, achieving 99% accuracy, precision, F1-Score, 98% recall, and 100% ROC AUC. This study offers a promising approach to early heart disease diagnosis and preventive healthcare.

## 1 Introduction

Heart disease includes a number of disorders that affect the heart. Heart disorders include conditions such as blood vessel disease, namely coronary artery disease. Cardiac arrhythmias Congenital heart defects refer to cardiac disorders that are present from birth [1]. Every year, 17.9 million people die from cardiovascular illnesses (CVDs) [2]. Coronary heart disease, cerebrovascular disease, rheumatic heart disease, and other illnesses are among the categories of heart and blood vessel disorders known as CVDs. Heart attacks and strokes account for more than four out of five CVD fatalities, and one-third of these deaths occur under the age of 70 [2]. Heart disease and stroke are the primary causes of CVDs, which kill 23.6 million people by 2030 [3].

WHO acknowledges that strokes are responsible for a significant number of deaths globally. Strokes cause most deaths and disabilities in low- and middle-income nations[4]. Healthy habits and better healthcare are essential to preventing strokes [5]. The main risk factors for heart disease in high-income nations are high blood pressure, high cholesterol, smoking, obesity, a lack of activity, poor diet, excessive alcohol use, diabetes, family history, age and gender, stress, and environmental factors [6]. On the other hand, long-term illnesses like heart disease are rising in nations with low or middle incomes. Changing diets, sedentary lifestyles, smoking, and poor healthcare access contribute to this global health concern[7]. Early stroke detection can save mortality in 85% of instances [5].

Healthcare is now cost-effective because of technology. Each artificial intelligence algorithm improves healthcare automation and disease prediction. Massive amounts of information are successfully analyzed and stored in the cloud using Hadoop, a system built on computer clusters [8] [9] [10]. Many applications, like text recognition and identification, early forecasting [11] [12], power quality disturbance detection[13], vehicle categorization, and agriculture, focus significantly on machine learning. When using modern tools to anticipate and diagnose cardiac illnesses, physicians tend to depend on their assessments of a patient's medical history, symptoms, and outcomes from physical examinations. Patient data, primarily medical records, is easily accessible and constantly growing in today's healthcare sector databases. To developing a prediction model for heart disease, this research analyzed the Stroke Prediction Dataset. The dataset's existing characteristics have been used to train the machine to identify patterns.

Bhatt et al,. [6] RF, DT, MP, and XGBoost. GridSearchCV hyper tuned model parameters to maximize results. DT: 86.37% (with cross-validation) and 86.53% (without), XGBoost: 86.87% and 87.02%, RF: 87.05% and 86.92%, multilayer perceptron: 87.28% and 86.94%, trained on 80:20 data split. This study found that multilayer perception with cross-validation is most accurate. Its greatest accuracy was 87.28%. In this study, the accuracy is not up to the mark. Erdoğan and Güney [14] proposed a technique for calculating the weight coefficient. The findings of the suggested technique show that 13 distinct patient characteristics led to an SVM: 86,90% success rate.

Given that heart disease is a serious health issue, it is essential to inform people about its risks and ways to prevent it. Men are more likely than women to get heart disease, according to statistics from Harvard Health Publishing [15], and this gap persists even after taking into account known risk factors. Several variables may cause a gender difference in the risk of getting heart disease, including: lifestyle decisions, biological and genetic factors, knowledge of the issue and seeking treatment, as well as changes in hormone levels [16].

In this century, machine learning has become accepted as a method for assessing data pertaining to illnesses. This paper will focus on the most efficient method for employing machine learning techniques in the early identification of heart disease in order to provide a solution.

The following actions have been taken to conduct more simulations:

- Processing the data to make the dataset prepared for analysis.
- Utilization of data preparation methods to improve and clean the dataset.

- Nine alternative models have been used for analyzing the dataset.
- Performance evaluation of different models to determine the one that performs the best.
- Assessment of the performance of the best model in comparison to previous research.

The research is divided into the following sections: introduction, related work, methodology, data collection, data conversion, and data preprocessing, applied models, result analysis, discussion, comparison with previous research, summary, and future scope. The section that follows offers a little more explanation in light of this.

## 2 Literature Review

Bharti *et al.*, [17] This study examines the results of the UCI Machine Learning Heart Disease dataset by using both machine learning and deep learning techniques. The confusion matrix demonstrates encouraging outcomes. The isolation forest technique is used to improve results by normalizing the data and reducing undesired attributes. The study also investigates the compatibility of multimedia technologies, including mobile devices. The deep learning model achieved a precision rate of 94.2%.

Bizimana *et al.*, [18] Using data scaling techniques, split ratios, ideal parameters, and machine learning algorithms, this study presents a machine learning-based prediction model (MLbPM) for heart disease (HD) prediction. In order to assess the proposed notion, experiments on an HD dataset from the University of California, Irvine demonstrate HD existence or absence. Using LR, resilient scaler, optimum parameter, and 70:30 split ratio, the proposed MLbPM achieves 96.7% accuracy. MLbPM is more accurate than earlier attempts as well.

Ch Anwar Ul Hassan *et al.*, [19] Eleven ML classifiers have been employed in this work to identify important predictors of heart disease. The prediction model was introduced using feature combinations and popular classification techniques. This work used gradient boosted trees and multilayer perceptron to predict heart disease with 95% accuracy. RF was used in this study with 96% accuracy.

Gupta and Raheja [20] NB, LR, DT Classifier, KNN, ABR, XGBoost, and RF Classifier are all used in this study. Following a thorough analysis of the data, it was discovered that the ABR, XGBoost, and RF Classifiers had the lowest percentages of inaccurate predictions and the highest accuracy ratings, with 95%, 96%, and 97%, respectively.

K. Karthick et al., [21]In this study, the author emphasizes the Cleveland HD dataset to develop a heart disease risk prediction model. SVM, GNB, LR, LightGBM, XGBoost, and RF algorithms have been used, and the accuracy was 80.32%, 78.68%, 80.32%, 77.04%, 73.77%, and 88.5%, respectively. The tests' results show that the RF method achieves 88.5% accuracy during validation for 303 data instances with 13 chosen Cleveland HD dataset attributes.

Shah et al., [22] The objective of the research was to apply machine learning to develop an algorithm model for predicting heart disease. The 303 patients and 17 features of the Cleveland heart disease dataset, which was taken from the UCI machine learning repository, have been used to generate the data for this paper. Several supervised classification techniques have been used by the authors, including naive Bayes, DT, RF, and KNN. The study's findings showed that, with 90.8% accuracy, the KKN model had the best level of precision. The study emphasizes the potential use of machine learning methods for predicting cardiovascular illness and stresses the significance of using the right models and methods to get the best outcomes.

Alotalibi [23] decided to take a look at the benefit of machine learning (ML) methods for finding heart disease. To develop prediction models, the study used a dataset from the Cleveland Clinic Foundation and used many ML methods, including DT, LR, RF, naive Bayes, and SVM. During the model design stage, a 10-fold cross-validation method was applied. The findings showed that the DT algorithm, which had a rate of 93.19%, and the SVM method, which had a rate of 92.30%, had the best accuracy in predicting heart disease. This work emphasizes the DT algorithm as a promising alternative for further research and sheds light on the potential of machine learning methods as an efficient tool for predicting heart failure disease.

G Rajkumar et al., [24] This study presents an enhanced deep learning-based framework utilizing the Hungarian heart disease dataset collected from IoT sensors. The approach includes data preprocessing using the median studentized residual method, feature selection via Harris Hawk Optimization, and disease classification using Modified Deep Long Short-Term Memory (MDLSTM) with modification through Improved Spotted Hyena optimization. The results showcase superior predictive accuracy of 98.01% and a significantly reduced error rate of 91.11%, surpassing existing techniques.

Li et al., [25] This study developed an interpretable model to predict heart failure patient mortality in ICUs using the SHAP-explained XGBoost model. The XGBoost model demonstrated the highest predictive performance (AUC: 0.824) compared to other machine learning models. The SHAP method revealed the top predictors, with blood urea nitrogen being the most influential. The interpretable model aids physicians in mortality risk prediction, enabling tailored treatment plans and optimal resource allocation.

Shumaila Shehzadi et al., [26] This study achieving a highly precise model for early detection is crucial to reducing fatalities. Previous attempts to enhance detection accuracy utilized various models and attributes, yielding suboptimal results. This research proposes an artificial approach to categorize the current stage of heart disease, utilizing machine learning methods such as LR, NB, and RF. Experimental outcomes demonstrate exceptional accuracy, with RF at 99%, NB at 97%, and LR at 98%. This high accuracy suggests a potential reduction in annual deaths due to ischemic heart disease.

# 3 Methodology

Finalization of this section required the completion of eight separate primary parts. In (Fig 1) shown the work flowchart of this study. The contents of the dataset's description are presented and discussed in one section of the "Data Collection" section. The history of the data set has also been thoroughly examined. The "Data Conversion" part transforms the string data into numerical values. The required data pretreatment methods have been modified in the report's section on "Data Preprocessing." The part labeled "Important Feature Selection" also uses eight different models to identify which critical characteristics are most beneficial. The dataset has been divided into a train set and a test set in the section of the text titled "Train Test Split" so that the experiment may be conducted on each of them separately. The study's "Applied Model" section includes a list of all nine models that have been used to assess the dataset and paper the likelihood of acquiring early stokes. "Result Analysis" Section The results of the model that performed the best have been scrutinized and compared with those of earlier published work to successfully predict the onset of stroke using machine learning.

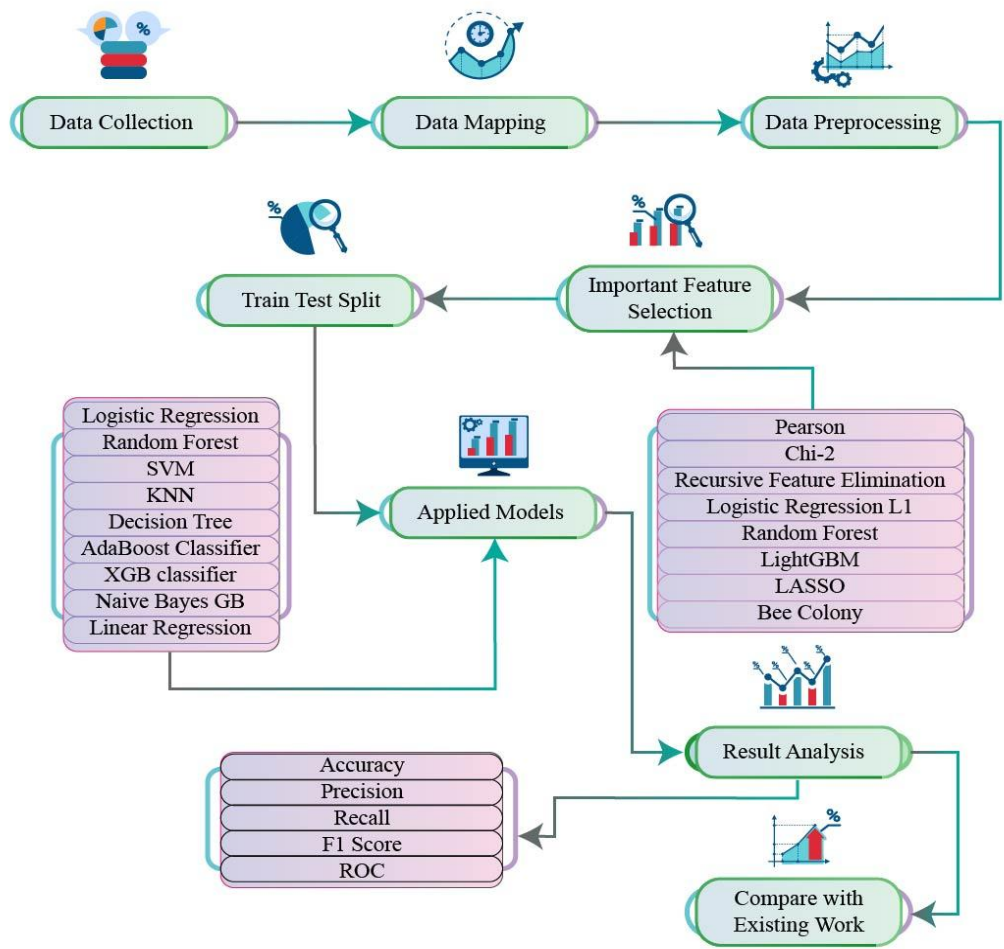

**Fig 1. System architecture of this study.**

The dataset section is currently starting the working phase of this study to the subsequent discussion of the dataset's characteristics can take place.

## 3.1 Dataset

The signs and symptoms of heart disease in patients who have recently been diagnosed or who are at risk of getting the condition are described in this dataset. 5110 observations with 12 characteristics make up the data. There is a dataset called Kaggle's Stroke Prediction Dataset. The dataset contains eleven clinical traits that can be used to predict specific stroke occurrences. In addition to demographic information like gender, age, marital status, and type of residence, these characteristics also contain health-related information like high blood pressure, heart disease, average blood sugar levels, body mass index (BMI), and smoking habits. Additionally, the collection contains occupational data that categorizes various types of employment. Individual tracking within the dataset is made possible by the unique identifier (Id). Whether a patient has suffered a stroke is specified by the stroke target variable (1). This dataset is a great resource for developing prediction models to assess stroke risk based on these several variables.

## 3.2 Data Attribute

Dataset attributes represent the characteristics utilized by the system, and for heart disease data, these include attributes such as the patient's gender, age, BMI, hypertension and others, as shown in Table 1.

**Table 1. The Dataset's Summarized Description**

| Attribute | Data Types | Description | |
|---|---|---|---|
| **Id** | Integer | Unique identifier for each patient | Valid : 100% Mismatched: 0% Missing: 0% |
| **Gender** | Object | Patient's gender (male, female, other) | Valid: 100% Mismatched: 0% Missing: 0% Unique: 3 Most Common: 59% (Female) |
| **Age** | Float | Age of the Patient | Valid: 100% Mismatched: 00% Missing: 00% Mean: 43.2 Std. Deviation: 22.6 |
| **Hypertension** | Integer | 0 if the patient doesn't have hypertension, 1 if the patient has hypertension | Valid: 100% Mismatched: 0% Missing: 0% Mean: 0.1 Std. Deviation: 0.3 |
| **Heart Disease** | Integer | 0 if the patient doesn't have any heart diseases, 1 if the patient has a heart disease | Valid: 100% Mismatched: 0% Missing: 0% Mean: 0.05 Std. Deviation: 0.23 |
| **Ever Married** | Object | No or Yes | Valid: 100% Mismatched: 0% Missing: 0% Yes: 3353 (66%) False: 1757 (34%) |
| **Work Type** | Object | Children, Govt. job, Never Worked, Private or Self employed | Valid: 100% Mismatched: 0% Missing: 0% Unique: 5 Most Common: 57% (Private) |
| **Residence Type** | Object | Rural or Urban | Valid: 100% Mismatched: 0% |

| | | | Missing: 0%<br>Unique: 2<br>Most Common: 51% (Urban) |
|---|---|---|---|
| **Average Glucose Level** | Numerical (float) | Average glucose level in blood | Valid: 100%<br>Mismatched: 0%<br>Missing: 0%<br>Mean: 106<br>Std. Deviation: 45.3<br>Min: 55.1<br>Max: 272 |
| **BMI** | Numerical (float) | Body mass index | Valid: 100%<br>Mismatched: 0%<br>Missing: 0%<br>Unique: 419<br>Most Common: 4% (N/A) |
| **Smoking Status** | Object | formerly smoked, never smoked, smokes or Unknown (unknown means that the information is unavailable for this patient) | Valid: 100%<br>Mismatched: 0%<br>Missing: 0%<br>Unique: 4<br>Most Common: 37% (Never smoked) |
| **Stroke** | Integer | 1 if the patient had a stroke or 0 if not | Valid: 100%<br>Mismatched: 0%<br>Missing: 0%<br>Mean: 0.05<br>Std. Deviation: 0.22<br>Min: 0<br>Max: 1 |

This dataset consists of 5110 patients and 12 features, including five categorical and seven continuous variables. The continuous property most likely corresponds to numerical health data, whereas the categorical attributes most likely suggest the demographics of the patient, medical records, or therapeutic types. Examining these traits can provide crucial details for medical comprehension and targeted stroke management action. Investigating the relationships between these traits might aid in the creation of individualized treatment strategies and improve patient results.

## 3.3 Data Preprocessing

In this part, data preprocessing methods have been employed in the following manner in preparation for the subsequent simulation that would predict the early onset of stroke. Separating the target (stroke) feature from the rest of the 11 characteristics, and storing them. Furthermore, data normalization has been applied for the continued feature "Age". There are a lot of outliers in avg_glucose_level and bmi. The outliers make the distribution curve of both features highly skewed towards the right. Either the outliers can be removed or the distribution curve can be made less-skewed by mapping the values with a log, but both cases will lead to loss of the number of data points with Stroke = 1, avg_glucose_level increases with age and similarly leads to more chances of stroke. The stroke class is highly imbalanced. There are 212 outliers in the avg_glucose_level that have a stroke=1, which include a large quantity of data. Removing them will lead to loss of data. Rather, remove the outliers in BMI because the number of outliers with strokes is just 10, which won't affect the dataset much.

## 3.4 Null Handling

Null handling, also known as missing data handling, is a crucial aspect of data preprocessing and analysis [27]. When dealing with real-world datasets, it's common to encounter missing or null values due to various reasons such as data entry errors, sensor malfunctions, or incomplete records. Efficient null handling involves strategies to either impute or remove these missing values to ensure robust and accurate analysis [28].

Let's denote a dataset $D$ with $n$ observations and $m$ features. The presence of null values can be expressed mathematically as:

$$D = \{x_{ij} | i = 1, 2, \ldots n; j = 1, 2, \ldots, m\} \tag{1}$$

Where $x_{ij}$ represents the value of the feature $j$ in observation $i$. If a value is missing, it is represented as a null or NaN: $x_{ij}$ = null. Null handling involves various strategies, including:

Removing observations or features with null values. Mathematically, this might be represented as:

$$D' = \{x_{ij} | x_{ij} \neq null\} \tag{2}$$

Filling in missing values with estimated or calculated values. For instance, mean imputation for a feature $j$ might be represented as:

$$\bar{x}_j = \frac{1}{n'} \sum_{i=1}^{n'} x_{ij}, \text{ Where } x_{ij} \neq null \tag{3}$$

$$x_{ij} = \begin{cases} \bar{x}_j, & if\ x_{ij} = null \\ x_{ij}, & if\ x_{ij} \neq null \end{cases}, \text{ Where } x_{ij} \neq null \tag{4}$$

Null handling is essential for maintaining the integrity of the dataset and ensuring that the analysis and modeling processes are not adversely affected by missing information. The code for handling null values is "df['bmi'].fillna(df['bmi'].mean(), inplace = True)" . In this dataset, bmi has 201 null columns. To simplify the calculation procedure, this study replaced the null values with '0'.

## 3.5 Outliers Handling

In the realm of artificial intelligence and machine learning, outliers act as misfits in a sea of data, often lurking as irregular data points that can lead models astray. Properly handling these anomalies is essential to ensuring the robustness and reliability of our algorithms [29]. Techniques like imputation, transformation, and the application of specialized models help us tame these data irregularities, allowing our AI systems to discern meaningful patterns and insights from the noise. By addressing outliers thoughtfully, we equip our models with the precision and resilience needed to make well-informed decisions, ultimately paving the way for more accurate predictions and valuable insights in diverse fields, from finance to healthcare [30].

Outliers can be represented as values that fall outside a specified range or deviate substantially from the mean or median of the dataset.

Let's define a dataset $D$ with $n$ observations:

$$D = \{x_1, x_2, \ldots, x_n\} \tag{5}$$

**where $x_i$ represents an individual data point.**

The average glucose level has 212 outliers with a stroke value of 1, which include a substantial amount of data. Removing them will lead to loss of data.

data.drop(data[data['bmi'] > 45].index, inplace = True)

data.drop(data[data['bmi'] < 12.7].index, inplace = True)

This study wants to remove the outliers in the bmi variable, since the number of outliers associated with stroke is just 10, which would have not much effect on the dataset. If bmi > 45 and bmi < 12.7, this study replaces values with 0.

## 3.6 Duplicate Checking

Duplicate checking involves identifying and managing repeated or identical entries within a dataset. Mathematically, duplicates can be expressed as observations or instances that have the same values across all features or a subset of features.

Consider a dataset $D$ with $n$ observations and $m$ features:

$$D = \{x_{ij} | i = 1, 2, \ldots, n; j = 1, 2, \ldots, m\} \tag{6}$$

Where $x_{ij}$ represents the value of the feature $j$ in observation $i$.

Duplicate checking involves identifying observations that are identical across all features or specific subsets. This can be represented as:

$$\textit{Exact Duplicate Checking}, x_i = x_j \textit{ for all } i \neq j \tag{7}$$

This condition indicates that the entire observations $x_i$ and $x_j$ are the same for all features.

$$\textit{Subset} - \textit{based Duplicate Checking}, x_{ik} = x_{jk} \textit{ for a specific subset of features } k \tag{8}$$

Here, only a subset of features $k$ is compared for equality between observations $x_i$ and $x_j$.

Identifying and managing duplicates is crucial to maintain data quality and prevent skewing statistical analysis or machine learning models due to redundant information. The dataset analysis found no duplicate entries, ensuring accuracy and reliability for future study.

## 3.7 Category Data Encoding

During the preprocessing stage, the scikit-learn package has been used to apply label encoding, which converted categorical variables into numerical representations. This conversion is essential to provide interoperability with machine learning models that need numerical inputs. The scikit-learn's LabelEncoder has been used to systematically encode categorical characteristics, hence improving the dataset's appropriateness for further model training and analysis. After applying label encoding, the resultant dataset has been examined to verify its modified structure and data types, verifying the successful transformation of categorical variables into numerical representation. This preprocessing stage establishes the groundwork for using a wide variety of machine learning algorithms in the analysis, enabling significant insights and predictions.

## 3.8 Histogram of attributes

Histograms are essential tools in statistical analysis, offering a graphical depiction of the dataset's distribution. Histograms represent the frequency of values falling inside particular intervals by using the mathematical principle called the Central Limit Theorem. Histograms provide an excellent visualization of the probability distribution by partitioning the data into bins and determining the frequency of observations inside each bin. This visual depiction facilitates the identification of trends, anomalies, and the general configuration of the data distribution. Histograms are helpful tools for analyzing the underlying properties of distinct variables in a dataset by using statistical ideas. The mathematical equation for a histogram can be expressed as follows:

Let $X$ be a random variable representing the dataset, and $x_1, x_2, x_3, \ldots \ldots \ldots . x_n$ be its values. Define a set of bins $B = [b_1, b_2, b_3, \ldots \ldots \ldots . b_n]$ such that $b_1 < b_2 < b_3 < \cdots \ldots \ldots . b_k$. The histogram $H(x)$ is a function that assigns frequencies to each bin:

$$H(x_i) = frequency\ of\ (x_i)\ in\ the\ dataset. \tag{9}$$

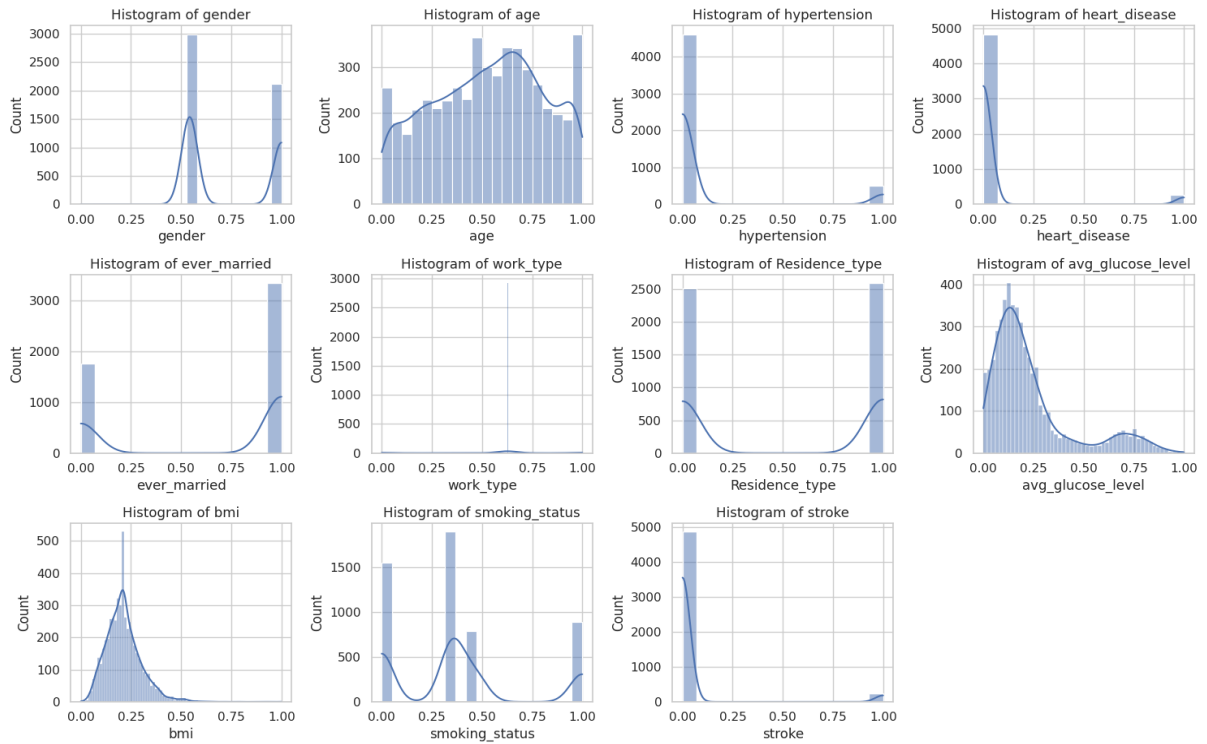

**Fig 2. Histogram of attributes**

The histograms in the (Fig 2) provide valuable insights into the distribution of key features used for stroke prediction. Starting with gender, the dataset consists of 59% females and 41% males, indicating a slight skew towards female patients. This balance is important when analyzing gender-related risk factors. The age distribution distances from 0.08 to 82 years, with a higher concentration of patients in middle age. This suggests that the dataset covers a wide range of ages, although it is more focused on adults and elderly individuals. In terms of hypertension, most patients do not have hypertension, as indicated by the vertical peak at 0 in the histogram. Only a small percentage of patients are hypertensive (value 1). Similarly, the heart disease histogram shows that most patients do not suffer from heart disease, with a small proportion having a positive history of heart conditions. The ever-married feature reveals that 66% of patients have been married, while 34% have not. This demographic feature could be relevant in lifestyle-related risk analysis. The work type histogram highlights that 57% of patients are employed in private jobs, 16% are self-employed, and 27% fall into other categories. This distribution reflects the variability in professional engagement among patients. In terms of residence type, the dataset is nearly evenly split, with 51% of patients residing in urban areas and 49% in rural settings. This balance ensures that the model can generalize well across different living environments. The average glucose level in the blood ranges from 55.1 to 272, with a notable concentration of patients having glucose levels between 55 and 150. This distribution emphasizes the potential role of glucose regulation in stroke prediction. The BMI histogram shows a typical distribution, with most patients having a BMI between 20 and 40, placing them in the healthy to slightly overweight range. Smoking status reveals diversity in the dataset, with a substantial number of non-smokers, alongside patients who have quit smoking. This variability is crucial for understanding how lifestyle choices impact stroke risk. Finally, the stroke histogram shows that most patients did not experience a stroke (value 0), with only a small fraction having had a stroke (value 1). This imbalance highlights the challenge of dealing with class imbalance in the dataset when training models for stroke prediction.

Overall, these histograms provide a clear representation of the distribution of the dataset's key features, highlighting crucial factors such as age, gender, medical history, and lifestyle, all of which are important for developing an effective stroke prediction model.

## 4 Feature Selection

The process of reducing the number of input variables while creating a predictive model is known as feature selection. Feature selection approaches are essential in the machine learning process as they improve model performance and reduce computational overhead. In the field of machine learning, the process of identifying relevant features is a crucial step in creating an ideal collection of features. The selection method entails identifying distinctive and influential traits, therefore reducing duplication in the feature space and overcoming the difficulties associated with the 'dimensionality curse' [31]. In certain situations, it is preferable to increase model performance while simultaneously lowering the computing burden of the model by decreasing the number of input variables.

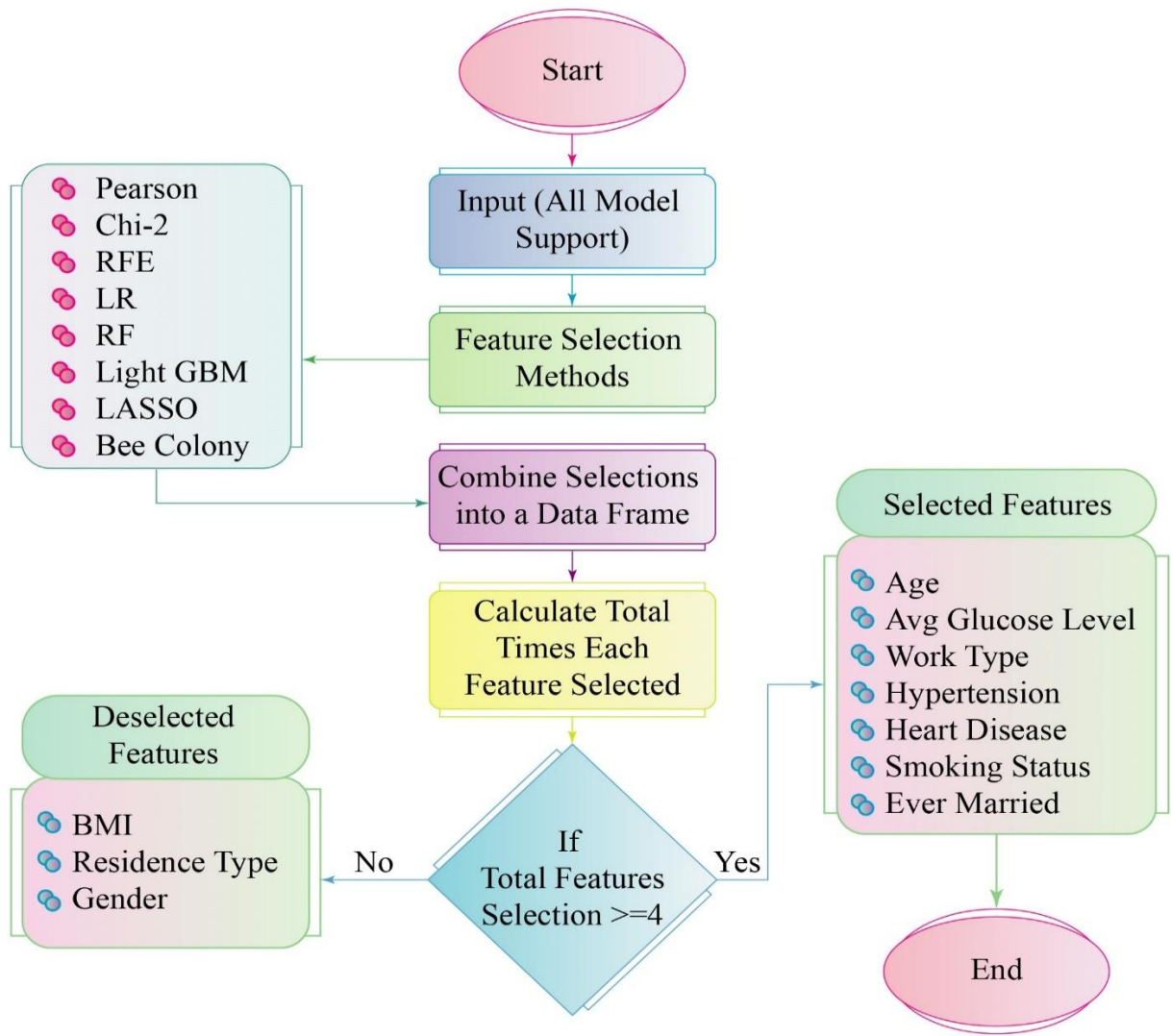

**Fig 3: Feature selection procedure**

The flowchart in (Fig 3) illustrates the process of feature selection for the classification of stroke disease using different models of machine learning and selection techniques. All model-supported features first enter the process, after which they are subjected to an array of feature selection techniques, such as LightGBM, LASSO, Pearson correlation, Chi-square (Chi-2), Recursive Feature Elimination (RFE), Logistic Regression (LR), Random Forest (RF), and Bee Colony optimization. Following the combination of the selections made using these approaches into a data frame. The convergence is determined by the stabilization of total number of times each feature is selected. A feature is kept as selected if total features selection >=4; if not, it is deselected. The model's selected features, as shown by the output, are age, work type, average blood sugar level, heart disease, hypertension, smoking status, and marital status (ever married). Features that didn't fit the selection criteria, such as gender, residence type, and BMI, were deselected. This flowchart clearly shows the methodical process of feature set improvement, which makes sure that only the best predictive variables are kept for model training. This improves the model's performance and readability for the classification of stroke diseases.

## 4.1 Features Correlation

Feature correlation, the interplay between variables, is the compass in the maze of data analysis. It unveils relationships, indicating shared influence or independence among features. High correlation hints at synchronization, potentially revealing redundancy or critical interdependence, while low correlation signifies divergence, offering distinct insights.

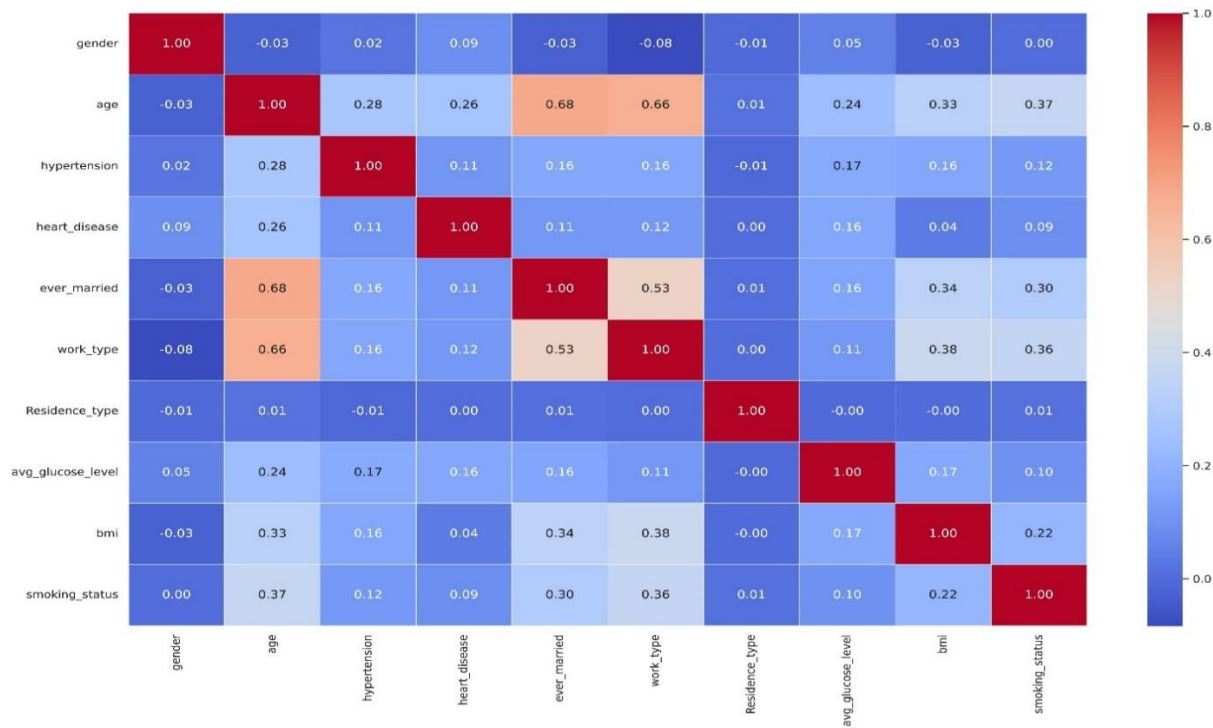

**Fig 4. Illustration of all Features Correlation**

In (Fig 4), we present an illustration depicting the correlation between all features used in this study. Understanding these correlations empowers the data scientist to sculpt a dataset that harmonizes interwoven features, fortifying the model with depth and accuracy.

### 4.1.1 Pearson

The test statistic that measures the statistical connection, or association, between two continuous variables is known as Pearson's correlation coefficient. The information provided includes the direction of the connection as well as the significance of the relationship or correlation. Two categories of correlations exist. In a positive correlation, when feature A rises, feature B rises as well, or when feature A falls, feature B falls as well. The two aspects have a linear connection and move in tandem. When there is a negative correlation, feature B reduces while feature A increases, and vice versa.

The formula for Pearson correlation between variables $X$ and $Y$ in a dataset with $n$ observations is:

$$r = \frac{\sum_{i=1}^{n}(X_i-\bar{X})(Y_i-\bar{Y})}{\sqrt{\sum_{i=1}^{n}(X_i-\bar{X})^2} \times \sqrt{\sum_{i=1}^{n}(Y_i-\bar{Y})^2}} \qquad (10)$$

Where, $X_i$ and $Y_i$ are individual data points for variables $X$ and $Y$. $\bar{X}$ and $\bar{Y}$ are the means of variables $X$ and $Y$, respectively.

The Pearson correlation coefficient signifies: r = 1: A perfect positive linear relationship; r = −1: A perfect negative linear relationship; r = 0: No linear relationship between the variables.

### 4.1.2 Chi-2

The chi-2 test is a type of statistical analysis that allows you to compare an investigation's actual results to what was anticipated. This test will assess whether a disparity between actual and predicted data can be attributed to random variation or whether it can be attributed to a relationship between the variables that are the subject of the research. The k value is set to 10 for the selected dataset. This test's goal is to establish whether the relationship between the variables that are the subject of the study and the discrepancy can be determined.

In a contingency table with categorical variables A and B, the observed frequencies are represented as $O_{ij}$, and the expected frequencies under the assumption of independence are denoted as $E_{ij}$,. The Chi-square statistic is calculated as:

$$x^2 = \sum \frac{(O_{ij}-E_{ij})^2}{E_{ij}} \qquad (11)$$

Where: $O_{ij}$ represents the observed frequency in cell $i$ and $j$, $E_{ij}$ is the expected frequency in cell $i$ and $j$.

The Chi-square test is widely applied in various fields to determine whether there is a statistically significant relationship between categorical variables, aiding in understanding associations or dependencies between different categorical factors in a dataset.

### 4.1.3 Recursive Feature Elimination (RFE)

Recursive Feature Elimination (RFE) is a feature selection method wherein subsets of characteristics are repeatedly chosen, and models are constructed to identify the most pertinent features. All characteristics are first added, a model is fitted, the features are ranked according to significance, and the least significant features are removed. In order to maximize model performance and decrease dimensionality, this process is continued until the required amount of features is obtained.

RFE is the common abbreviation for recursive feature elimination. In this study, the parameters are set as follows: estimator = LogisticRegression(), n_features_to_select = 8, step = 10, and verbose = 5.

### 4.1.4 Logistic Regression L1 (LR L1)

Important features are chosen using the LR with L1 normalization (Lasso) feature selection method, also known as LR with L1 regularization (LR L1). L1 regularized LR is presently regarded as standard practice in the field of machine learning. This approach may be used for a variety of classification problems, especially those involving a large number of unique features. By adding a penalty term depending on the absolute values of the coefficients, L1 regularization promotes sparsity in this method by pushing some feature coefficients to zero. Features that have coefficients that are not zero are deemed significant and are chosen for the model. When working with high-dimensional datasets, this can be very helpful in determining the most pertinent characteristics. It is imperative to solve a convex optimization problem before using L1 regularized LR. As a result, the parameters are adjusted at threshold = 1.25 × median and penalty = l2.

The LR model predicts the probability of a binary outcome (0 or 1) based on input features. The L1 regularization term is added to the standard LR cost function, typically known as the binary cross-entropy or log loss function.

The LR L1 model's cost function is:

$$J(w) = \frac{1}{m} \sum_{i=1}^{m} \left( -y^{(i)} \log(\hat{y}^{(i)}) - (1 - y^{(i)}) \log(1 - \hat{y}^{(i)}) \right) + \lambda ||\omega||1 \quad (12)$$

Where, $J(w)$ is the cost function with regularization. $m$ is the number of training samples. $y^{(i)}$ represents the actual class label of the $i-$th sample. $\hat{y}^{(i)}$ represents the predicted probability of the $i-$th sample belonging to the positive class. $\omega$ is the vector of model weights. $\lambda$ is the regularization parameter, controlling the strength of regularization. $\lambda||\omega||1$ denotes the L1 norm of the weight vector, which is the sum of the absolute values of the weights.

### 4.1.5 Random Forest (RF)

RF is an ensemble learning technique that selects features by evaluating the importance of each feature as the model is trained. Measuring a feature's contribution to reducing an impurity (such as the Gini impurity) during prediction-making establishes the feature's relevance in a RF model. Characteristics that lead to a greater decrease in impurity are deemed to be more important. Every DT in a RF, which can contain 400–2,000 DT, is constructed using a random subset of the characteristics and observations from the dataset. DT in a RF might number from 400 to 2,000. There can be 400–12,000 DT in a RF at any given time. As a result, the parameters are set to threshold = 1.25 × median and n_estimators = 100.

### 4.1.6 LightGBM

A gradient boosting framework called LightGBM gives priority to teaching with the tree-based learning methodology. Unlike other methods, the LightGBM generates trees vertically. The majority of tree-growing algorithms create trees in a horizontal manner. This implies that the LightGBM methodology builds trees leaf-wise instead of level-wise, in contrast to previous approaches. In other words, the settings are set at min_child_weight = 40, reg_alpha = 3, reg_lambda = 1, num_leaves = 32, min_estimators = 500, learning_rate = 0.05, and colsample_bytree = 0.2.

### 4.1.7 LASSO

LASSO, also known as Least Absolute Shrinkage and Selection Operator, is a widely used regularization approach in the fields of linear regression and machine learning. The penalty term is included in the basic linear regression goal function to encourage the coefficients of less significant characteristics to approach zero. The regularization term is directly proportional to the magnitudes of the coefficients, and the intensity of regularization is determined by a hyperparameter known as the regularization parameter (alpha). The L1 norm penalty applied to the linear regression objective function defines the optimization problem for Lasso regularization. Standard linear regression minimizes the sum of squared residuals:

$$minimize \sum_{i=1}^{n}(y_i - \hat{y}_i)^2 \qquad (13)$$

The objective function is modified in Lasso regularization by adding a penalty term proportionate to the sum of the absolute coefficient values multiplied by a regularization parameter $\alpha$:

$$minimize \sum_{i=1}^{n}(y_i - \hat{y}_i)^2 + \alpha \sum_{j=1}^{p}|w_j| \qquad (14)$$

In this case, $y_i$ stands for the response variable that was observed, $\hat{y}_i$ for the response variable that was predicted, $w_j$ for the regression coefficients, and $\alpha$ for the degree of regularization.

In this study, LASSO is applied for feature selection using the SelectFromModel method with a specified alpha (alpha = 0.001) parameter. The training set is transformed with the selected features, followed by a similar transformation for the test set. The workflow continues with modeling or analysis using the transformed sets. The support and feature names after LASSO are obtained, and the code prints the number of selected features. This visual representation illustrates the sequential steps involved in the feature selection process using LASSO.

### 4.1.8 Bee Colony

Bee Colony Optimization (BCO) is a bio-inspired algorithm that emulates the foraging activity of honeybees to carry out feature selection. BCO seeks to maximize a fitness function $f(s)$ in mathematical terms, which represents the quality of a chosen subset of features $(s)$. Bee Colony Optimization iteratively explores and exploits solutions, updating employed bees based on fitness evaluations. The best solution is continuously updated throughout the process. The output reveals the selected features after BCO, providing an optimized feature set for enhanced model performance. This visual representation illustrates the comprehensive steps involved in combining advanced feature selection techniques with optimization algorithms to identify an optimal feature subset for machine learning models. The summarized output of the eight model for the important feature selection has been shown in table 2.

**Table 2. The Summarized Output of the Eight Model for the Important Feature Selection**

| SL | Feature | Pearson | Chi-2 | RFE | Logistics | RF | Light GBM | LASSO | Bee Colony | Total |
|---|---|---|---|---|---|---|---|---|---|---|
| 1 | Age | True | True | True | True | True | True | True | True | 8 |
| 2 | Avg Glucose Level | True | True | True | True | True | True | True | False | 7 |
| 3 | Work Type | True | True | True | True | False | False | True | True | 6 |
| 4 | Hypertension | True | True | True | True | False | False | True | False | 5 |
| 5 | Heart Disease | True | True | True | True | False | False | True | False | 5 |
| 6 | Smoking Status | True | True | True | False | True | False | False | False | 4 |
| 7 | Ever Married | True | True | True | False | False | False | True | False | 4 |
| 8 | BMI | True | False | False | False | True | True | False | False | 3 |
| 9 | Residence Type | True | True | True | False | False | False | False | False | 3 |
| 10 | Gender | True | False | False | False | False | False | False | False | 1 |

The most important features for predicting stroke were selected from Table 2. Specifically, the top seven components with a total value of four or above were chosen. These attributes will be used to forecast whether an individual will have a stroke or not. The variables include age, average glucose level, job type, hypertension, heart disease, smoking status, and marital status.

## 4.2 Train Test Split

The nine important features that have been selected have been divided into a train set and a test set in order to carry out the application of the models and complete the task of stroke prediction. This meant that 80% of the dataset was utilized for training and just 20% was used for testing.

**Table 3. Description of the Train Test Split Dataset**

| Name | Description |
|---|---|
| **Percentage of train set** | 80% |
| **Percentage of the test set** | 20% |
| **Number of patients overall** | 5110 |

In this study, the dataset used for stroke prediction consisted of 5,110 patients. In Table 3 the data was split into two sets: 80% for training and 20% for testing. This division was essential to ensure that the models were properly trained on most of the data while retaining a portion for evaluating their performance. The train-test split allowed for model validation, ensuring that the models could generalize well to unseen data. The training set consisted of approximately 4,088 patients, while the test set included 1,022 patients, providing a balanced framework for model evaluation and stroke prediction.

## 4.3 K-Fold Cross Validation

One method for evaluating a machine learning model's resilience and performance is K-fold cross-validation. The initial dataset is divided into K equal-sized "folds" or subgroups. K-1 folds are used to train the model, while the remaining fold is used to validate it. Each fold is utilized as the validation data precisely once during the K iterations of this method.

**K- Fold Cross Validation's Key Steps:**
1. There are K folds, or subgroups, inside the dataset.
2. The remaining fold is used to verify the model once it has been trained on (K-1) folds.
3. For every fold, performance indicators (such accuracy and loss) are noted.
4. To give an overall evaluation of the model's performance, the average performance over all K folds is computed.

K-Fold Cross Validation makes sure that each data point is utilized for validation precisely once, which contributes to a more reliable assessment of the model. It's helpful in determining any problems, like overfitting or underfitting, as well as how the model will generalize to new data.

$$\textit{Cross-validation accuracy} = \frac{1}{k} \sum_{i=1}^{k} Accuracy_i \quad (15)$$

Where: k is the number of folds in the cross-validation, $Accuracy_i$ represents the accuracy of the model on the $i$th validation set or fold.

This equation sums up the accuracy obtained on each validation set and divides it by the total number of folds to calculate the average accuracy across all the validation sets.

## 4.4 Applied Models

### 4.4.1 XGBoost Classifier (XGB)

The open-source program XGBoost is a popular and efficient implementation of the gradient boosted tree method. Gradient boosting in supervised learning combines predictions from less complex, less dependable models to anticipate a target variable with accuracy. Regression trees, in which each leaf maintains a continuous score for each input data point, are used as weak learners for regression problems. XGBoost uses a convex loss function based on the difference between expected and target outputs, together with a penalty for model complexity, to minimize a regularized objective function (L1 and L2 regularization). Iteratively adding new trees that forecast residuals or mistakes from prior trees is what happens during training. These fresh trees along with the older ones make up the final forecast. The method adds new models and optimizes by minimizing information loss. Objective (reg:linear), colsample_bytree (0.3), learning_rate (0.1), max_depth (5), alpha (10), and n_estimators (10) are important XGBoost settings.

The XGBoost Classifier algorithm demonstrated an accuracy of 95% when applied to the Cardiovascular Disease Dataset (Kaggle) by Bhatt et al. [6]. It achieved a 79.80% accuracy rate when applied to the UCI Dataset by Erdoğan and Güney [14], and a 96% accuracy rate by Gupta and Raheja [20]. Additionally, it achieved a 73.77% accuracy rate when applied to the UCI ML repository's Cleveland heart disease dataset by K. Karthick et al. [21].

### 4.4.2 Random Forest (RF)

A method of classification consisting of several individual DT is referred to as a RF. In order to produce an ensemble of uncorrelated trees with a higher prediction accuracy by committee than any single tree, bagging and feature the randomization are employed. Different sets of characteristics and data are used to construct each tree. The strategy aims to improve forecast precision. 'gini' as the splitting criteria and 100 as the number of estimators (trees) are two of the RF's important parameters [32].

$$\text{RF Prediction} = \frac{1}{i} \sum_{i=1}^{i} p_i(x) \tag{16}$$

where, $p_i(x)$ is the mean of the predictions produced by each of the $i$ regression trees, and $i$ is the number of independent regression trees generated for the bootstrap samples using the input vector $x$.

The RF algorithm achieved the following accuracies with different datasets: 95% with the Cardiovascular Disease Dataset (Kaggle) by Bhatt et al. [6], 96% with the UCI-Repository dataset by Ch Anwar Ul Hassan et al., [19], 98.6% with the Cardiovascular Disease Dataset by AbdElminaam et al., [33], 85.15% with the UCI Dataset by Erdoğan and Güney [14], 88.52% with the UCI to Predict the heart illness dataset by Farzana and Veeraiah [34], 97% by Gupta and Raheja [20], and 88.5% with the UCI ML repository's Cleveland heart disease dataset by K. Karthick et al. [21].

### 4.4.3 K-Nearest Neighbor (KNN)

KNN is a Machine Learning algorithm used for classification and regression tasks. In KNN, a data point's classification or value is determined by the majority vote or averaging of its K closest neighbors in the feature space, based on a defined distance metric. KNN is a non-parametric algorithm and is relatively simple to implement, making it a popular choice for various applications. However, it may be sensitive to the choice of K and the distance metric used[35].
The KNN algorithm achieved an accuracy of 82.10% with UCI Dataset by Erdoğan and Güney [14], an accuracy of 67.21% with UCI to Predict the heart illness Dataset by Farzana and Veeraiah [34].

### 4.4.4 Support Vector Machine (SVM)

One popular and adaptable supervised machine learning method is the SVM. Activities requiring regression and classification may both be finished with its help. The classification job, however, will be the topic of discussion in this thread. It is generally thought to be best for small and medium-sized data sets. The main objective of the SVM is to find the optimal hyperplane that maximizes the margin and splits the data points into two components linearly[36]. For optimal performance, the random state is set to 0, and the kernel is set to "linear."

The SVM algorithm achieved an accuracy of 83.50% with UCI Dataset by Erdoğan and Güney [14], an accuracy of 80.32% with UCI ML repository's Cleveland heart disease dataset by K. Karthick et al., [21], an accuracy of 81.97% with UCI to Predict the heart illness Dataset by Farzana and Veeraiah [34].

### 4.4.5 Adaptive Boosting Regression (ABR)

Using a sequential ensemble approach, ABR creates a strong and dependable learner by combining many beginners that are randomly selected from the dataset. These ineffectual learners are generated utilizing a variety of machine learning techniques. Weights are assigned to each sample observations throughout each training iteration, influencing the learning process of each hypothesis.

This approach identifies and assigns more relevance to instances that have produced inaccurate predictions, giving them precedence for further training with a new base learner. The repeated method continues until the algorithm achieves accurate classification. Within the scope of regression analysis, the conclusion of an instance is not characterized as correct or erroneous, but rather denotes an absolute value miscalculation, typically a constant. The ensemble prediction is achieved by computing the median or weighted average of the predictions given by each individual base learner. This technique enhances the model's ability to handle complex information and improve forecast accuracy [37], [38]. The ABR algorithm achieved an accuracy of 95% by Gupta and Raheja [20]

### 4.4.6 Gaussian Naïve Bayes (GNB)

GNB is known as supervised algorithm. Using the Gaussian Naive Bayes algorithm, a probabilistic classification technique is shown. The Bayes theorem and strong independence assumptions form the foundation of this approach. Assuming that the continuous values associated with each class are distributed according to a normal (or Gaussian) distribution is a common method when dealing with continuous data. This is done in order to make dealing with continuous data easier. We'll continue based on the following assumption about the possibility of the qualities: In the Gaussian Naive Bayes technique, continuous valued features and models are thought to individually correspond to a Gaussian distribution (also known as a normal distribution). It is capable of using the Naïve Bayes model and does not use Bayesian approaches. Naive Bayes classifiers are used in several intricate real-world scenarios[39], [40].

$$F(A/B) = \frac{F(B/A) \times F(A)}{F(B)} \tag{17}$$

$F(A/B)$ is the posterior probability, $F(A)$ is the class prior probability, $F(B)$ is the predictor prior probability, $F(B/A)$ is the likelihood, probability of predictor.

The Naïve Bayes algorithm achieved an accuracy of 84.90% with UCI Dataset by Erdoğan and Güney [14], 78.68% accuracy with UCI ML repository's Cleveland heart disease dataset by K. Karthick et al., [21], an accuracy 82.25% with UCI to Predict the heart illness Dataset by Farzana and Veeraiah [34].

### 4.4.7 Logistic Regression (LR)

LR is the appropriate regression approach for use when a dependent variable is dichotomous. Like other regression examinations, the LR is a predictive research. To sum up the data and explain the association between one dependent binary variable and one or more independent nominal, ordinal, interval, or ratio-level variables, this study use LR. where the penalty is set to 12 and the random_state is set to zero.

The LR algorithm achieved an accuracy of 96.7% with Cleveland UCI HD dataset by Bizimana *et al.*, [18], an accuracy of 84.77% with UCI Dataset by Erdoğan and Güney [14], 80.32% accuracy with UCI ML repository's Cleveland heart disease dataset by K. Karthick et al., [21]

### 4.4.8 Linear Regression

A data analysis method called linear regression uses a related, known data value to predict the value of unknown data. We used linear regression to explore the relationships between features and the outcome, which helped us validate the relevance of the selected features and ensured that even simple models could capture essential patterns in the data. Using a linear equation, it represents the unknown, or dependent, and the known, or independent, variables quantitatively. It's like trying to figure out which straight line best fits this connection. Reducing the discrepancy between the actual target values and the anticipated values from the line is the aim. To put it simply, linear regression calculates the intercept and coefficients that characterize this straight line. These factors aid in calculating the relative effect on the goal of each modification in each attribute. The procedure entails using an approach known as Ordinary Least Squares to minimize the sum of squared discrepancies between the actual target values and the projections[41].

The linear regression is represented by a generic equation.

$$\boldsymbol{p = r + qz} \tag{18}$$

The values of r and q are the p-intercepts and slope, respectively. The equation represents the line that provides the most accurate fit to the data.

$$\boldsymbol{p = qz} \tag{19}$$

Statistically, this equation is frequently expressed as

$$\boldsymbol{p = \beta_0 + \beta_{1a1}} \tag{20}$$

The general equation is expressed as follows when there are $n$ predictors $(a_1, a_2, \ldots\ldots, a_n)$.

$$\boldsymbol{p = \beta_0 + \beta_{1a1} + \beta_{2a2} + \beta_{3an}} \tag{21}$$

### 4.4.9 Decision Tree Classifier

A DT is a type of decision support tool that represents options and the expected outcomes of those actions using a model resembling a tree. These possible consequences include, but are not limited to, resource prices, resource utility, and the outcomes of random occurrences. An algorithm that is nothing more complex than a collection of conditional control statements may be shown using this. In operations research, DT, or more accurately decision analysis, are a widely used method for determining which plan has the best chance of succeeding. In machine learning, DT are a widely used technique. 'gini' is the criteria that is used in this instance.

$$\boldsymbol{Entropy\ (Q) = \sum_{i=1}^{a} -Pi\ log_2\ Pi}, \tag{22}$$

$$\boldsymbol{Gain\ (Q,Y) = Entropy\ (Q) - \sum_{v \in Values\ (Y)} \frac{|Qv|}{|Q|} Entropy\ (Qv)} \tag{23}$$

The acquired findings are more comprehensible and understandable [42]. This approach has superior accuracy compared to other algorithms due to its analysis of the dataset using a tree-like graph structure. Nevertheless, the data could be excessively categorized and decision-making is conducted by testing just one characteristic at a time.

An accuracy of 94% with Cardiovascular Disease Dataset (Kaggle) has been achieved by the DT by Bhatt et al. [6], 78.20% accuracy with UCI Dataset by Erdoğan and Güney [14]

## 4.5 Performance Metrics

Performance metrics are measurable outcomes that are used to evaluate the way machine learning algorithms and models work. They provide a methodical approach to evaluate a model's performance for a specific job, including clustering, regression, classification, or other sorts of data analysis[43]. Four common performance metrics are included in this study: F1-Score (1-4), Accuracy, Precision, and Recall.

**Accuracy:** The proportion of accurately anticipated cases to all occurrences is known as accuracy.

$$A = \frac{TP+TN}{TP+TN+FP+FN} \tag{24}$$

**Precision:** Precision may be defined as the percentage of all positive forecasts that are genuine positive predictions.

$$P = \frac{TP}{TP+FP} \tag{25}$$

**Recall:** The percentage of genuine positive forecasts among all real positives is known as recall (sensitivity).

$$R = \frac{TP}{TP+FN} \tag{26}$$

**F1-Score:** A balance between recall and accuracy, calculated as the harmonic mean of the two.

$$F1 - Score = \frac{TP}{TP+FP} \tag{27}$$

## 4.6 Best Model Selection

This research paper included a thorough investigation of machine learning model selection and hyperparameter optimization. The scope of this research includes a wide range of models given in `list_of_models`, with each model being created using a specific `create_model()` method. The purpose of this function is to generate and customize each model using a unique combination of parameters.

In order to enhance the efficiency of these models, this study used a systematic hyperparameter search using `GridSearchCV`, a reliable technique that thoroughly investigates all possible combinations of hyperparameters. The hyperparameter grid, referred to as `param_grid`, is customized for each model and includes parameters such as 'param1', 'param2', and other features relevant to the model. The grid search is performed using 5-fold cross-validation to assure reliability and avoid overfitting. Additionally, parallel processing is used to enhance computational efficiency (with `n_jobs=-1`).

After conducting a hyperparameter search, this research selected the best-performing model by identifying the optimum hyperparameters using the `grid_search.best_estimator_` method. In order to evaluate how well the model performs in different scenarios, this research used cross-validation using the resampled training data (`x_train_resampled` and `y_train_resampled`). This approach allowed for a comprehensive assessment of the model's prediction skills, as shown by the `cv_scores`.

Following that, this research used the chosen model to forecast results on the test data (`x_test`). The precision of these forecasts was then evaluated using the `accuracy_score` metric to assess the model's efficacy.

**Pseudocode**

```
// import necessary libraries

for each model in list_of_models:
        model = create_model()
        // Define the hyperparameter grid to search
        param_grid =
                {
                        'param1': [value1, value2, ...],
                        'param2': [value1, value2, ...],
                        # ... additional parameters
                }

        // Create GridSearchCV to find the best hyperparameters
        grid_search = GridSearchCV(estimator=model, param_grid=param_grid, cv=5,
        scoring='accuracy', n_jobs=-1)
        grid_search.fit(x_train_resampled, y_train_resampled)

        // Get the best model with the best hyperparameters
        best_model = grid_search.best_estimator_

        cv_scores = cross_val_score(best_model, x_train_resampled, y_train_resampled, cv=5,
        scoring='accuracy')
 end for

// Predict using Test Data
y_pred = best_model.predict(x_test)

// Calculate Accuracy
accuracy = accuracy_score(y_pred, y_test)

// Print Result
if grid_search.best_score_ is not None:
    print('Best Accuracy:', grid_search.best_score)
else:
   print('Best Accuracy not available.')

print('Test Accuracy:', accuracy)
print('Best Hyperparameters:', grid_search.best_params)
```

This research placed a high importance on openness and dependability while reporting these study results. The report provides detailed information on the outcomes, including the highest attainable accuracy after tweaking the hyperparameters, the accuracy on the test data, and the specific hyperparameters that resulted in optimum performance. The research included a conditional statement to account for scenarios when obtaining the highest level of accuracy may not be possible, therefore assuring the reliability of the results.

This study not only enhances the progress of machine learning applications, but also offers a systematic and transparent method for selecting models and fine-tuning hyperparameters, which can be used to other domains and datasets.

In our study, hyperparameter tuning was conducted using grid search across all machine learning models. For each model, we defined a comprehensive parameter grid containing key hyperparameters that significantly impact model performance. For example, the XGBoost model's grid included hyperparameters such as learning rate, max depth, subsample, and the number of estimators. Similarly, the grid for the Random Forest model encompassed the number of trees, maximum features, and maximum depth.

We employed GridSearchCV with 5-fold cross-validation to systematically evaluate and optimize these hyperparameters for each model. Performance metrics—accuracy, precision, recall, F1-score, and ROC AUC—were computed for each hyperparameter combination. The optimal hyperparameters were identified based on the best cross-validation score, ensuring that the models did not suffer from overfitting.

Among all the models, XGBoost emerged as the top performer, achieving the highest accuracy. The grid search successfully identified the best set of hyperparameters, and when these were applied to the test data, the model demonstrated outstanding accuracy. This underscores the effectiveness of XGBoost in predicting heart disease risk, indicating its strong potential for early diagnosis and intervention.

**Table 4. Hyperparameters of machine learning algorithms used grid search**

| Model | Hyperparameter | Definition | Value Range |
|---|---|---|---|
| **Linear Regression** | fit_intercept | to calculate the intercept in the model | [True, False] |
| | copy_X | to copy the input X before fitting | [True, False] |
| **LR** | penalty | Type of regularization (L1 = Lasso, L2 = Ridge) | ['l1', 'l2'] |
| | C | Inverse of regularization strength | [0.1, 1, 10] |
| **DT** | max_depth | Maximum depth of the tree | [None, 10, 20] |
| | min_samples_split | Minimum number of samples required to split a node | [2, 5, 10] |
| | min_samples_leaf | Minimum number of samples required to be at a leaf node | [1, 2, 4] |
| **ABR** | n_estimators | Number of weak learners (trees) used | [50, 100, 150] |
| | learning_rate | Contribution of each weak learner | [0.1, 0.01, 0.001] |
| **KNN** | n_neighbors | Number of neighbors to consult when predicting | [3, 5, 7, 9] |
| | weights | Weighting strategy for the neighbors | ['uniform', 'distance'] |
| | p | Distance metric: L1 (Manhattan) or L2 (Euclidean) | [1, 2] |
| **NB-Gaussian** | var_smoothing | Variance smoothing parameter to prevent numerical issues | [1e-9, 1e-8, 1e-7] |

|  | priors | Class priors (None = uniform, or specified) | [None, [0.2, 0.3, 0.5], [0.1, 0.4, 0.5]] |
|---|---|---|---|
| **SVM** | C | Regularization parameter balancing margin and classification error | [0.1, 1, 10, 100] |
|  | gamma | Kernel coefficient for non-linear classification | ['scale', 'auto', 0.1, 0.01, 0.001] |
|  | class_weight | Weighting of classes to handle imbalance | [None, 'balanced'] |
| **XGBoost** | n_estimators | Number of trees used | [50, 100, 150] |
|  | max_depth | Maximum depth of the trees | [3, 5, 7] |
|  | learning_rate | Shrinks contribution of each tree | [0.01, 0.1, 0.2] |
|  | min_child_weight | Minimum sum of instance weight (hessian) for child node splitting | [1, 5, 10] |
| **RF** | n_estimators | Number of trees in the forest | [50, 100, 150] |
|  | max_depth | Maximum depth of the trees | [None, 10, 20, 30] |
|  | min_samples_split | Minimum number of samples required to split a node | [2, 5, 10] |
|  | min_samples_leaf | Minimum number of samples required at a leaf node | [1, 2, 4] |

In this study, we tuned the hyperparameters of nine machine learning algorithms using grid search to optimize their performance for heart disease prediction. Key parameters such as regularization strength, tree depth, number of estimators, and distance metrics were adjusted for each model. XGBoost achieved the highest accuracy after fine-tuning. Table 4. Hyperparameters of machine learning algorithms used in grid search summarizes the key hyperparameters and their respective value ranges across all models.

The hyperparameters of various machine learning models were fine-tuned to enhance performance and control overfitting. For linear regression, tuning involved adjusting the fit_intercept (whether to calculate the intercept) and copy_X (whether to copy the input data), while for logistic regression, the penalty (L1 for Lasso, L2 for Ridge) and C (regularization strength) were optimized, allowing for better control over the bias-variance trade-off. Similarly, in DT classifiers, parameters such as max_depth (tree depth), min_samples_split (minimum samples to split a node), and min_samples_leaf (minimum samples for a leaf node) were adjusted to reduce overfitting. Furthermore, in ABR, the number of weak learners (n_estimators) and the learning rate were fine-tuned, striking a balance between learning speed and the potential for overfitting. For KNN, optimization was achieved by varying n_neighbors (number of neighbors), weights (uniform or distance-based), and p (L1 for Manhattan or L2 for Euclidean distance). Additionally, in Gaussian Naive Bayes, var_smoothing (to handle low variance) and priors (to control prediction bias) were tuned. Moreover, for SVM, performance was enhanced by adjusting C (regularization), gamma (kernel coefficient), and class_weight (for handling class imbalance). Similarly, XGBoost, known for its superior results, was fine-tuned using n_estimators (number of boosting rounds), max_depth, learning_rate (shrinkage of tree contribution), and min_child_weight (minimum weight for splitting nodes). Lastly, for RF, hyperparameters such as n_estimators (number of trees), max_depth, min_samples_split, and min_samples_leaf were optimized, collectively improving the model's overall accuracy and reducing overfitting risks. Through these adjustments, each model's predictive performance was significantly enhanced.

## 4.7 Confusion Matrix

An effective visual tool for evaluating an algorithm's effectiveness is the confusion matrix. It offers a succinct and straightforward method for looking at prediction mistakes. Four essential elements make up this matrix: true negatives (TNs), false positives (FPs), false negatives (FNs), and true positives (TPs). Actual class instances are arranged as rows in the matrix, while expected class instances are organized as columns[44]. The confusion matrix may include many metrics, including accuracy, recall, and F1 score, in addition to showing mistakes. Every one of these measures has a unique function and is used in certain situations.

# 5 Results Analysis

## 5.1 Cross Validation with Grid Search

Table 5 presents a comprehensive summary of the overall efficacy of several machine learning models. Each model performed cross-validation using grid search to improve hyperparameters. The table presents crucial performance metrics, such as accuracy, precision, recall, and F1 score, for each model. The XGBoost model had exceptional overall performance, with an accuracy score of 0.99. This indicates a substantial level of precision in the model's predictions. Furthermore, the XGBoost model consistently attains exceptional precision, recall, and F1 score values of 0.99, 0.98, and 0.99, respectively. The RF model has a high level of accuracy, namely 0.98. The model has exceptional prediction abilities, evident from its elevated precision, recall, and F1 score metrics, which stand at 0.97, 0.98, and 0.98, respectively. The KNN model has outstanding accuracy, with a score of 0.99. Furthermore, it consistently maintains a well-balanced Precision value of 0.98 and Recall value of 0.97, leading to an outstanding F1 Score of 0.98. The SVM achieved a commendable accuracy of 0.98, but its precision is noticeably lower at 0.89. SVM has an excellent recall rate of 0.97 and an overall satisfactory F1 score of 0.93. The ABR model achieves an accuracy of 0.80, a precision of 0.75, a recall of 0.89, and an F1 score of 0.81, which are all impressive performance metrics. While its performance may not be as outstanding as that of other models. The precision of the NBG is 0.89. The F1 score of 0.79 is obtained from the precision value of 0.76 and the recall value of 0.81. The logistic regression model attained an accuracy of 0.78, a precision of 0.74, a recall of 0.85, and an F1 score of 0.79, showing a very effective prediction power. The accuracy of the linear regression model is 0.77. These models provide comparable levels of precision, recall, and F1 score, with the DT model slightly outperforming linear regression.

**Table 5. Overall Performance of All of the Models Summed Up Cross Validation with Grid Search**

| Model | Accuracy | Precision | Recall | F1 Score |
|---|---|---|---|---|
| **XGBoost** | **0.99** | **0.99** | **0.98** | **0.99** |
| **RF** | 0.98 | 0.97 | 0.98 | 0.98 |
| **KNN** | 0.99 | 0.98 | 0.97 | 0.98 |
| **SVM** | 0.98 | 0.89 | 0.97 | 0.93 |
| **ABR** | 0.80 | 0.75 | 0.89 | 0.81 |
| **NB-Gaussian** | 0.89 | 0.76 | 0.81 | 0.79 |
| **LR** | 0.78 | 0.74 | 0.85 | 0.79 |
| **Linear Regression** | 0.77 | 0.73 | 0.84 | 0.78 |
| **DT** | 0.97 | 0.94 | 0.98 | 0.97 |

In summary, the results provide a thorough comprehension of the relative efficacy of each model, enabling educated decisions on model selection based on specific performance measures and objectives. The outstanding results shown by XGBoost and KNN highlight their suitability for circumstances that need a fine balance between accuracy and the constraints of accuracy-recall. The most effective methods discovered for the proposed models significantly prefer XGBoost, demonstrating an amazing accuracy of 99%, precision of 99%, recall of 98%, F1 score of 99%, and a Roc AUC of 100%.

In (Fig 5) illustrates a graphical representation of all models confusion matrix.

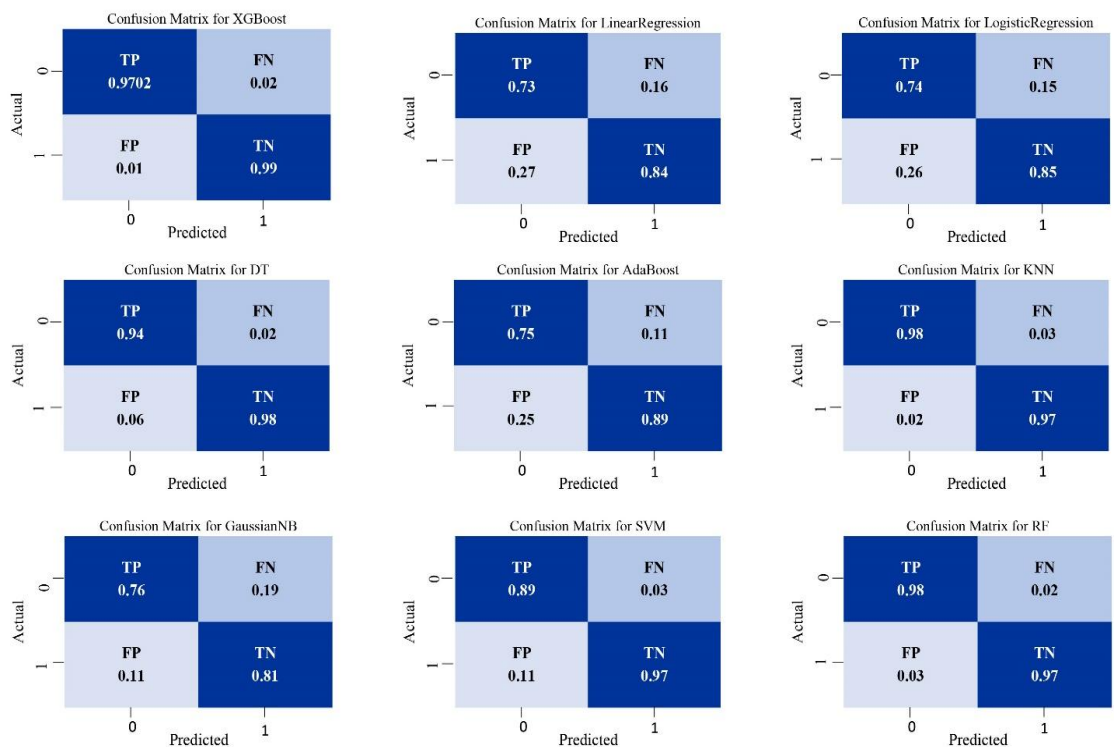

**Fig 5. Graphical Representation of All Models Confusion Matrix**

This research used confusion matrices to deeply look at and evaluate the prediction capabilities of different models. Evaluating the confusion matrices offers valuable information into the models performance across different classes. The XGBoost and RF algorithms show strong performance, characterized by a high level of sensitivity and a low level of specificity. This indicates their efficacy in reliably identifying positive situations. The DT method has outstanding accuracy in selecting positive situations while effectively reducing the frequency of false positives. However, SVM and ABR encounter challenges in reliably detecting positive occurrences, evident from their lower true positive rates and higher false positive rates. These findings contribute to the ongoing discourse on the process of choosing and refining models, with tangible implications for practical implementations in the real world. The comprehensive evaluation presented in the confusion matrices is a valuable tool for scholars and professionals seeking to deploy effective machine learning solutions across several domains.

## 5.2 ROC AUC

The Receiver Operating Characteristic Area Under the Curve (ROC AUC) scores are important for evaluating the ability of different machine learning models to identify between classes in binary classification tasks. Several models have been analyzed in this study and their prediction accuracy and precision have been assessed based on their individual ROC AUC values. The highest-performing models in this examination are XGBoost, KNN and RF both showing excellent ROC AUC values of 100%. This indicates that the models are almost perfectly able to recognize between the two classes in the binary classification task. The high ratings indicate that these models are both strong and accurate in their predictive capabilities, making them highly suitable for deployment in situations where accuracy is crucial. The model's high score demonstrates its strong capacity to accurately differentiate between the classes, making it a dependable option for the given position. The SVM model, while not achieving the same exceptional performance as XGBoost, KNN, RF nonetheless exhibits a commendable ROC AUC score of 93%. Strong discriminative capacity is shown by this score, indicating that the SVM model is a good fit for applications requiring precise classification.

Among the models with intermediate ROC AUC values are the ABR, NBG, LR, and linear regression. Their scores, which fall between 77% and 80%, show respectable predictive performance. Although these models do not reach the same degree of performance as the top-performing models, they nonetheless provide a reliable level of accuracy and precision in distinguishing between both binary classes. The DT Classifier has remarkable performance in class differentiation, as shown by its outstanding ROC AUC value of 97%. The performance of this model is remarkable, making it a very suitable choice for situations where precise categorization is of utmost importance. In (Fig 6) shows the ROC curve for the analysis of various models used in this study.

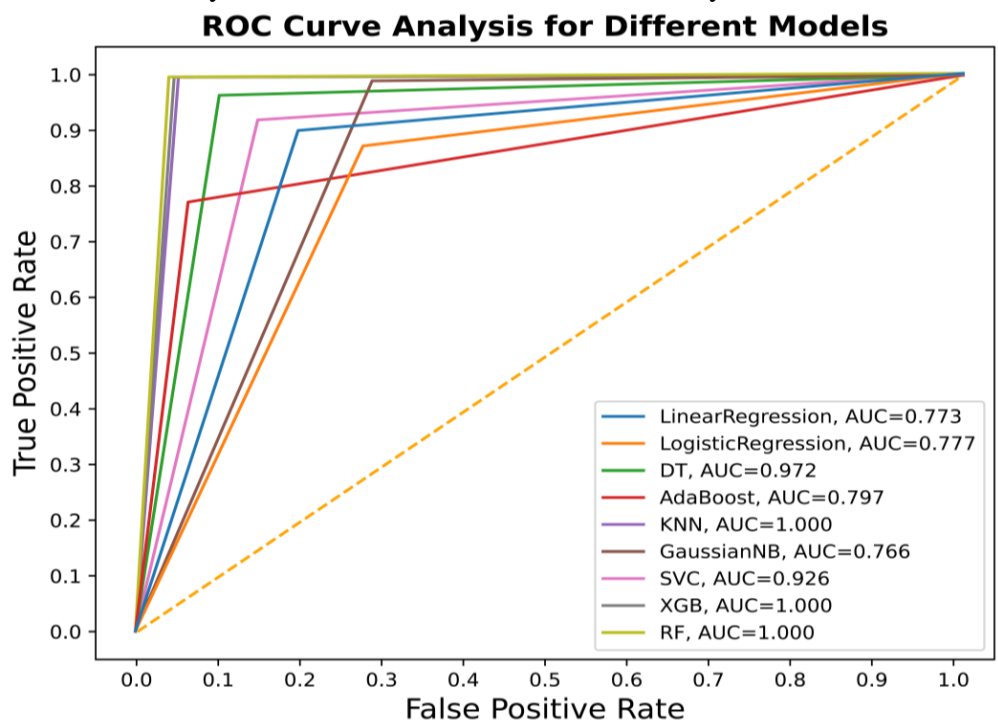

**Fig 6. The ROC Curve for the Experiment**

To summarize, ROC AUC ratings are crucial for evaluating and selecting models for accurate binary classification tasks. Models with higher scores, such as XGBoost, KNN, RF Classifier, and DT Classifier, have stronger discriminative ability and are likely to excel in reliably categorizing instances inside the binary classification issue. Thorough evaluation of these scores is crucial for making well-informed judgments on the use of machine learning models in real-world scenarios.

# 6 Ablation Study

## 6.1 Without Cross Validation and Grid Search

When machine learning models have been evaluated in Table 6 without the use of grid search or cross-validation, XGBoost and RF showed good accuracy (0.94) but had problems with recall and F1 score. The DT model exhibited a commendable accuracy of 0.92, but had challenges in terms of precision and recall. KNN and SVM emerged as the top performers, with both achieving an accuracy of 0.88. The KNN algorithm had an exceptional recall of 0.98, highlighting its ability to accurately identify positive cases. The SVM exhibited a balanced trade-off between accuracy and recall. The ABR and NB-Gaussian exhibited similar performance, achieving an accuracy of 0.78. LR and linear regression demonstrated lower accuracy levels of 0.74 and 0.73, respectively, while also displaying distinct precision-recall features. In general, KNN and SVM are particularly suitable for tasks that need a delicate trade-off between accuracy and recall, but XGBoost demonstrated outstanding performance across all measures.

**Table 6. Overall Performance of All of the Models Summed Up Without Cross Validation and Grid Search**

| Model | Accuracy | Precision | Recall | F1 Score | ROC |
|---|---|---|---|---|---|
| XGBoost | 0.94 | 0.25 | 0.03 | 0.06 | 0.51 |
| RF | 0.94 | 1.00 | 0.17 | 0.03 | 0.51 |
| **KNN** | **0.88** | **0.81** | **0.98** | **0.89** | **0.88** |
| SVM | 0.88 | 0.84 | 0.94 | 0.89 | 0.87 |
| ABR | 0.78 | 0.72 | 0.92 | 0.80 | 0.78 |
| NB-Gaussian | 0.78 | 0.77 | 0.83 | 0.79 | 0.78 |
| LR | 0.74 | 0.17 | 0.84 | 0.28 | 0.78 |
| Linear Regression | 0.73 | 0.16 | 0.81 | 0.27 | 0.77 |
| DT | 0.92 | 0.21 | 0.13 | 0.16 | 0.55 |

## 6.2 Cross Validation without Grid Search

In Table 7, utilizing cross-validation but without grid search, RF Classifier and XGBoost models maintained a high accuracy of 0.95, although their precision, recall, and F1 score varied. SVM achieved a balanced accuracy across all metrics with a slight dip in precision. KNN maintained a high recall, F1 score and ROC, but the precision was slightly lower. ABR had a moderate accuracy of 0.72, and Naive Bayes-Gaussian maintained consistency with the previous scenario.

**Table 7. Overall Performance of All of the Models Summed Up Cross Validation without Grid Search**

| Model | Accuracy | Precision | Recall | F1 Score | ROC |
|---|---|---|---|---|---|
| **XGBoost** | 0.95 | 0.25 | 0.03 | 0.05 | 0.51 |
| **RF** | 0.95 | 0.94 | 1.00 | 0.02 | 0.51 |
| **KNN** | 0.91 | 0.88 | 0.99 | 0.93 | 0.92 |
| **SVM** | **0.92** | **0.91** | **1.00** | **0.95** | **0.94** |
| **ABR** | 0.72 | 0.65 | 0.84 | 0.75 | 0.73 |
| **NB-Gaussian** | 0.78 | 0.77 | 0.83 | 0.79 | 0.78 |
| **LR** | 0.75 | 0.17 | 0.81 | 0.28 | 0.78 |
| **Linear Regression** | 0.79 | 0.16 | 0.81 | 0.27 | 0.77 |
| **DT** | 0.92 | 0.21 | 0.13 | 0.13 | 0.55 |

**Comparison of the performance of ML models for prediction of Cross Validation with Grid Search vs Without Cross Validation and Grid Search vs Cross Validation without Grid Search**

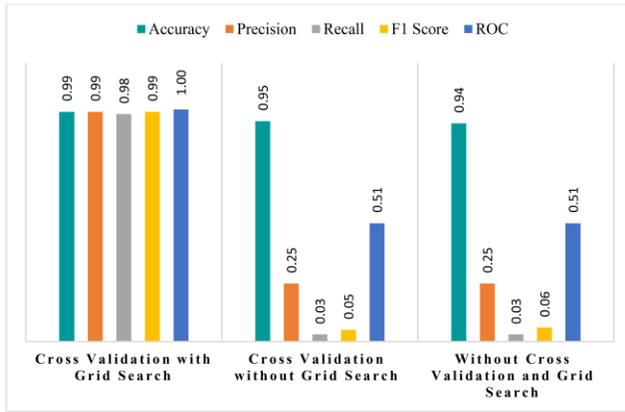
(a) XGB

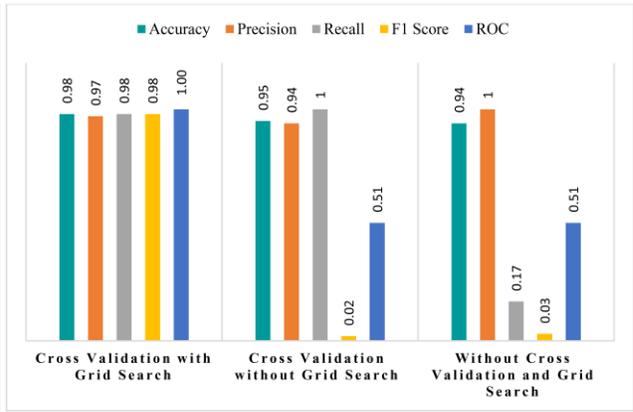
(b) RF

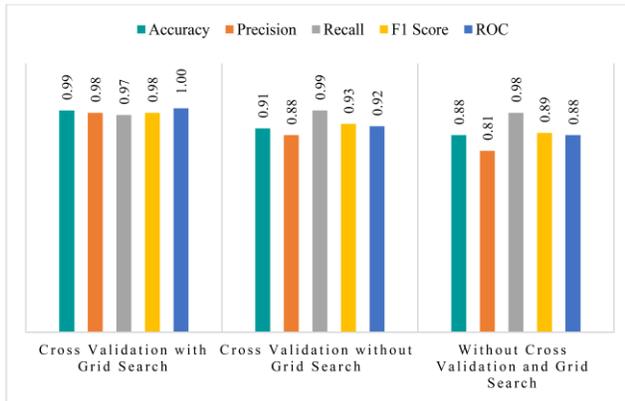
(c) KNN

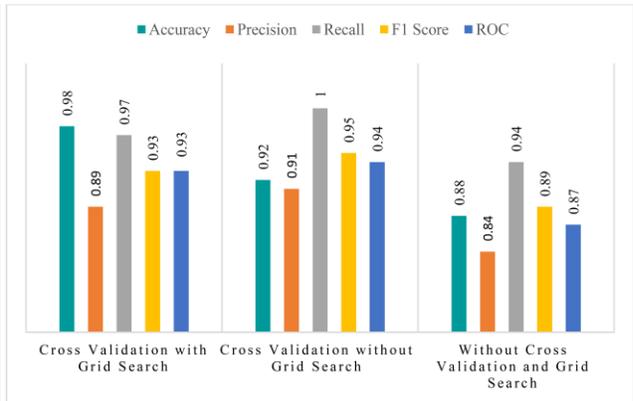
(d) SVM

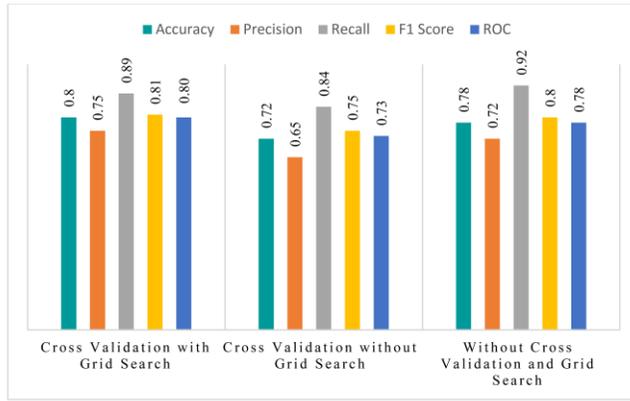

(e) ABR

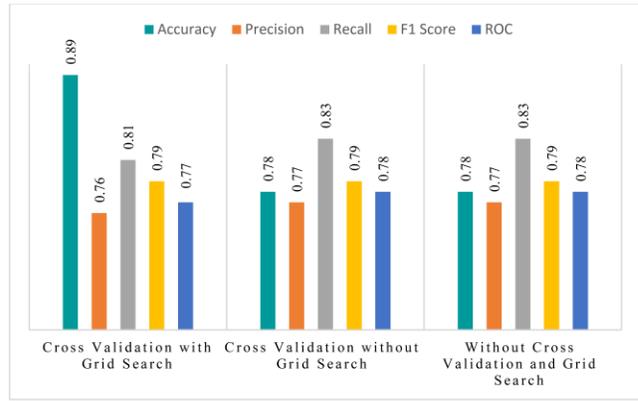

(f) NB-Gaussian

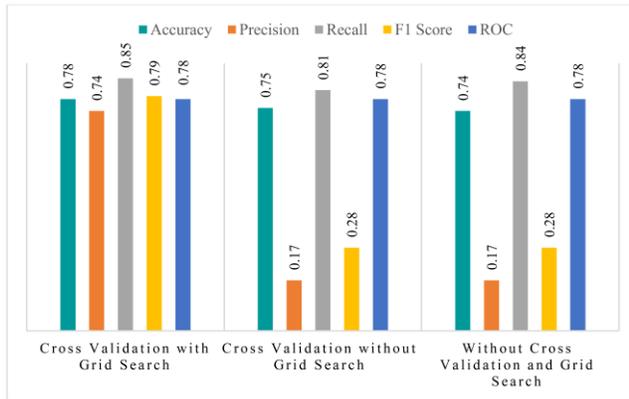

(g) LR

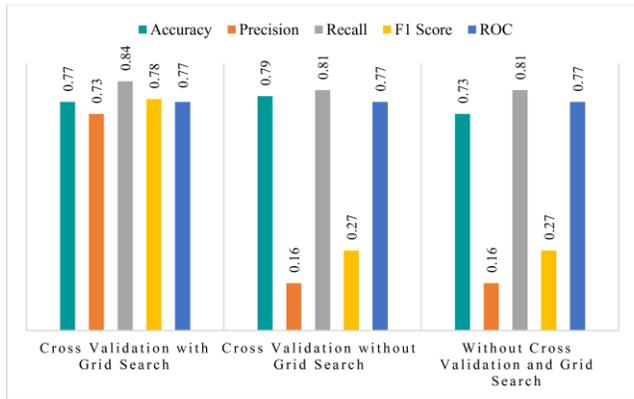

(h) Linear Regression

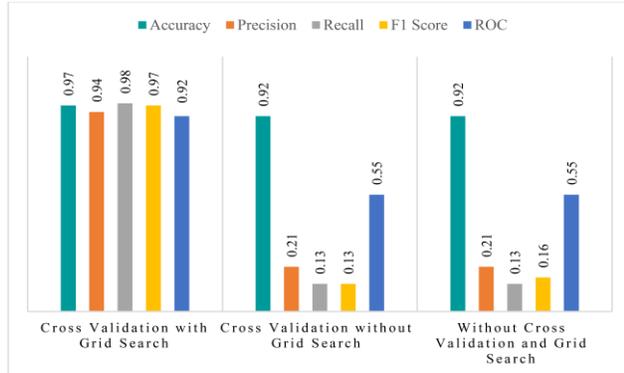

(i) DT

**Fig 7.** (a) XGBoost Model performance on Cross Validation with Grid, Cross Validation without Grid Search, Without Cross validation and Grid Search; (b) RF Model performance on Cross Validation with Grid, Cross Validation without Grid Search, Without Cross validation and Grid Search; (c) KNN Model performance on Cross Validation with Grid, Cross Validation without Grid Search, Without Cross validation and Grid Search; (d) SVM Model performance on Cross Validation with Grid, Cross Validation without Grid Search, Without Cross validation and Grid Search; (e) ABR Model performance on Cross Validation with Grid, Cross Validation without Grid Search, Without Cross validation and Grid Search; (f) NB-Gaussian Model performance on Cross Validation with Grid, Cross Validation without Grid Search, Without Cross validation and Grid Search; (g) LR Model performance on Cross Validation with Grid, Cross Validation without Grid Search, Without Cross validation and Grid Search; (h) Linear Regression Model performance on Cross Validation with Grid, Cross Validation without Grid Search, Without Cross validation and Grid Search; (i) DT Model performance

on Cross Validation with Grid, Cross Validation without Grid Search, Without Cross validation and Grid Search

In summary, Table 5, which incorporates cross-validation and grid search, is the most robust approach for model evaluation and selection, as it consistently yields the best-performing models. This analysis underscores the importance of employing these techniques in model development to ensure reliable and effective classification results, as well as the significance of tuning hyper-parameters through grid search for optimizing model performance.

# 7 Discussion

In this study, we enhanced the classification accuracy of stroke disease by combining several feature selection strategies with a voting system. This strategy is different from many previous research that usually use a random vote mechanism or depend on a single feature selection method. On the other hand, our method ensures that the most relevant features are selected by integrating techniques like LASSO, Recursive Feature Elimination (RFE), Pearson correlation, and others. This multi-method approach considers a greater range of feature selection, resulting in a more robust model. Additionally, our work examines the effectiveness of advanced models like XGBoost and LightGBM, which are known for their superior performance in classification tasks, while many prior studies concentrated on conventional machine learning models like Support Vector Machines (SVM) and Logistic Regression. Our study stands out from previous studies that frequently used default hyperparameters or less complex tuning techniques by utilizing these models in conjunction with grid search for hyperparameter optimization.

## 7.1 Comparison with Existing Works

In the context of healthcare, specifically in the field of cardiovascular disease and stroke prediction, various machine learning models and algorithms have been applied to different datasets. To better understand the effectiveness of these approaches, the study compares the performance of these models with existing work and their respective datasets. Here is an analysis of the results from several studies:

Bhatt et al. [6] This study employed RF, DT, and XGBoost models on the cardiovascular disease dataset. The reported accuracy scores are quite impressive, with all models achieving accuracy above 94%, showcasing their effectiveness in classifying cardiovascular disease cases. Bizimana et al. [18] In the analysis of the Cleveland UCI HD dataset using LR, the reported accuracy stands at 96.7%, indicating strong predictive capabilities for heart disease. Ch Anwar Ul Hassan et al. [19] The UCI-Repository dataset was utilized, and RF achieved an accuracy of 96%. This approach demonstrates the potential of RF in heart disease prediction. AbdElminaam et al. [33] Using the cardiovascular disease dataset and a RF model, an accuracy of 98.6% was achieved, signifying the robustness of this model in identifying cardiovascular disease cases.

Erdoğan and Güney [14] Multiple machine learning algorithms have been tested on the UCI dataset. Each algorithm achieved respectable accuracy scores ranging from 78% to 85%. The diversity of algorithms provides options for different scenarios and dataset characteristics. Farzana and Veeraiah [45] This study applied various models, including GNB, SVM, RF, KNN, and XGBoost to the UCI heart illness prediction dataset. The accuracy results vary, with RF reaching 88.52% accuracy, while K-Nearest Neighbour achieved 67.21%. The choice of model appears to significantly impact performance. Gupta and Raheja [20] ABR, XGBoost, and RF models have been explored, and all achieved high accuracy scores, with RF leading at 97%. This suggests the potential of ensemble methods in heart disease prediction. K. Karthick et al. [21] Several machine learning algorithms have been applied to the UCI ML repository's Cleveland heart disease dataset. The results ranged from 73.77% to 88.5% accuracy, emphasizing the importance of selecting the right algorithm for a specific dataset.

In the context of stroke prediction using the Stroke Prediction Dataset, various machine learning models have been employed. In this study, we achieved notably high accuracies across several models, with XGBoost and KNN both reaching 99%, showcasing the effectiveness of these models in predicting cardiovascular diseases.

**Table 8: The Summarized Comparison of the Heart Disease Prediction Model's Performance**

| Method | Dataset Name | Description | Accuracy |
|---|---|---|---|
| **Bhatt et al. [6]** | Cardiovascular Disease Dataset (kaggle) | RF | 95% |
| | | DT | 94% |
| | | XBG | 95% |
| **Bizimana *et al.*, [18]** | Cleveland UCI HD dataset | LR | 96.7% |
| **Ch Anwar Ul Hassan *et al.*, [19]** | UCI-Repository dataset | RF | 96% |
| **AbdElminaam *et al.*, [33]** | Cardiovascular Disease Dataset | RF | 98.6% |
| **Erdoğan and Güney [14]** | UCI Dataset | SVM | 83.50% |
| | | KNN | 82.10% |
| | | DT | 78.20% |
| | | Naive Bayes | 84.90% |
| | | LR | 84.77% |
| | | RF | 85.15% |
| | | XGBoost | 79.80% |
| | | LightGBM | 81.10% |
| **Farzana and Veeraiah [34]** | UCI to Predict the heart illness. | GNB | 82.25% |
| | | SVM | 81.97% |
| | | RF | 88.52% |
| | | KNN | 67.21% |
| | | XGB | 78.69% |
| **Gupta and Raheja [20]** | | ABR | 95% |
| | | XGBoost | 96% |
| | | RF | 97% |
| **K. Karthick et al., [21]** | UCI ML repository's Cleveland heart disease dataset | SVM | 80.32 |
| | | NB-Gaussian | 78.68% |
| | | LR | 80.32% |
| | | LightGBM | 77.04% |

|  |  | XGBoost | 73.77% |
|  |  | RF | 88.5% |
| **This Study** | Stroke Prediction Dataset | **XGBoost** | **99%** |
|  |  | **RF** | **98%** |
|  |  | **KNN** | **99%** |
|  |  | **SVM** | **98%** |
|  |  | **ABR** | **80%** |
|  |  | **NB-Gaussian** | **89%** |
|  |  | **LR** | **78%** |
|  |  | **Linear Regression** | **77%** |
|  |  | **DT** | **97%** |

Table 8 presents an analysis of previous research in the field of predicting stroke and heart disease. It shows the advantages and disadvantages of various machine learning models as well as how well they perform on various datasets. It emphasizes that ensemble methods such as RF and boosting algorithms like XGBoost tend to perform well. Moreover, the selection of an appropriate dataset and model diversity are crucial factors in achieving accurate predictions in this domain. The study has shown remarkable efficacy, providing vital insights for future research and practical applications in the healthcare field.

# 8 Clinical Integration and Implementation

This study proposes a cloud-based application designed to integrate advanced machine learning models into clinical workflows, specifically for stroke prediction. The application simplifies data administration, improves diagnostic precision, and fosters provider collaboration. (Fig 8) shows a scenario of a proposed clinical application.

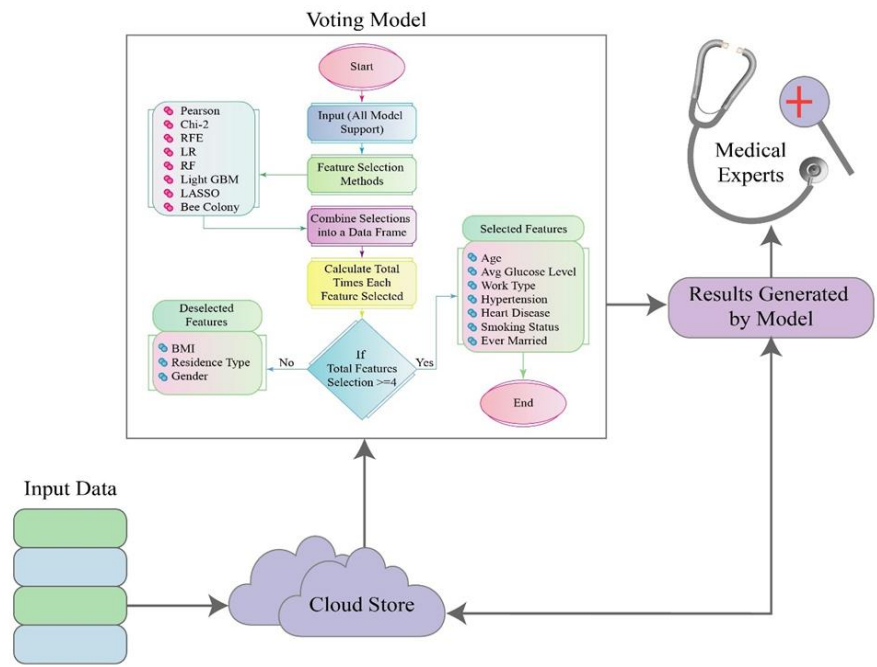

**Fig 8.** The illustration of proposed clinical application.

**Data Upload and Secure Cloud Storage:** Healthcare providers can upload patient data, including demographic information, clinical measurements, and lifestyle factors, directly into the application. The data is securely stored in a cloud environment, ensuring both data integrity and accessibility for further analysis. The centralized cloud storage system enables seamless data management and ensures that patient information is readily available for processing.

**Integration with Predictive Models and Feature Selection:** The application includes several feature selection techniques, such as LASSO, Bee Colony Optimization, Random Forest, Random Forest, Pearson correlation, Chi-square testing, Recursive Feature Elimination (RFE), Logistic Regression, and LightGBM. These techniques assess the relative importance of various attributes, which are subsequently combined via a voting mechanism. To make sure that only the most important predictors are included in the models for prediction, the most relevant attributes are chosen based on how frequently they occur across various techniques.

**Automated Feature Selection and Classification:** The program uses trained machine learning models to automatically process patient data after identifying the appropriate characteristics. These algorithms evaluate the data and categorize patients according to their risk levels; they have been optimized for stroke prediction. The models offer clinicians useful diagnostic support by properly predicting the risk of stroke.

**Collaborative Decision-Making and Real-Time Result Communication:** The healthcare professionals receive real-time updates on the outcomes produced by the prediction models. By enabling physicians to remark, discuss, and share the results within the platform, the tool promotes teamwork. This makes it possible for interdisciplinary consultations, in which several medical specialists work together to provide patient care, resulting in more precise and knowledgeable therapeutic decisions.

**Data Privacy and Security:** The application is designed with stringent data protection measures, including encryption and robust access controls. These measures ensure that patient data remains confidential and secure throughout the entire process, from data upload to analysis and storage.

**Integration with Existing Healthcare Systems:** Using a single, unified interface, healthcare providers can access patient data, medical history, and prediction results due to the application's seamless integration with Electronic Health Records (EHR) systems. This integration ensures that all pertinent data is available while making clinical choices and optimizes the workflow, cutting down on the time needed to obtain and analyze patient data.

# 9 Conclusion

In this study, we focused on predicting heart diseases by leveraging nine machine learning models and tuning hyperparameters through grid search. The objective was to explore the predictive capabilities of various classifiers and assess their accuracy in identifying stroke risk factors. Additionally, we enhanced heart disease classification by integrating feature selection techniques with a novel voting system to improve the predictive accuracy of machine learning models. Out of all nine models, XGBoost stood out with remarkable results, reaching 99% in accuracy, precision, and F1-Score, 98% in recall, and 100% in ROC AUC. The findings revealed that the combination of feature selection and the voting-based classification model significantly improved prediction accuracy. Practically, this approach has important implications for early detection of heart diseases in healthcare, while theoretically, it highlights the importance of optimizing features and ensemble methods for better outcomes. The results strongly supported the hypothesis that the integrated approach would enhance prediction accuracy. This research emphasizes the potential for data-driven healthcare applications and demonstrates how such techniques can be applied to life-saving tasks. Reflecting on the process, it emphasized the value of interdisciplinary methods in solving complex problems. However, the study's dataset limitations, including a restricted study population, as well as a lack of diversity in ethnicities and geographic locations, may affect the generalizability of the results to broader populations with diverse demographic and clinical characteristics. In future research, we plan to address these limitations by applying the model in real-world clinical environments, refining the voting system, and expanding the dataset to include more diverse populations for improved generalizability.

# 10   Data Availability Statement

The datasets for this study can be found in this link: https://www.kaggle.com/datasets/fedesoriano/stroke-prediction-dataset [46].

# 11   Acknowledgements

Thank you to all of the authors who made significant contributions to this paper.

# 12   Conflict of Interest statement

To the best of our knowledge, no financial or other conflict of interest exists.